\documentclass{article}

\usepackage{arxiv}

\usepackage[utf8]{inputenc} 
\usepackage[T1]{fontenc}    
\usepackage[hypertexnames=false]{hyperref}       
\usepackage{bookmark}
\usepackage{xurl}           
\usepackage{url}            
\usepackage{amsfonts}       
\usepackage{nicefrac}       
\usepackage{microtype}      
\usepackage{graphicx}
\usepackage{amsmath}
\usepackage[numbers]{natbib}

\usepackage{booktabs}             
\usepackage{multirow}             
\usepackage{makecell}             
\usepackage{tabularx}             

\usepackage{algorithm}
\usepackage{algorithmicx}
\usepackage{algpseudocode}
\usepackage{listings}

\usepackage{xcolor}
\usepackage{float}
\usepackage{enumitem}
\usepackage{placeins}

\makeatletter
\providecommand*{\toclevel@algorithm}{0}
\makeatother

\graphicspath{ {./images/} }

\title{SelKV: Selective KV Cache Merging with Per-Token Merge-or-Drop \\ and Attention Compensation}

\author{
 Soumia Bouyahiaoui\thanks{Corresponding author}  \\
National School of Artificial Intelligence (ENSIA), \\
Sidi Abdellah Campus,  Algiers, Algeria\\
  \texttt{soumia.bouyahiaoui@ensia.edu.dz} \\
   \And
 Manel Kara laouar \\
  National School of Artificial Intelligence (ENSIA), \\
  Sidi Abdellah Campus,  Algiers, Algeria\\
\texttt{manel.karalaouar@ensia.edu.dz} \\
  \And
 Aicha Boutorh \\
  National School of Artificial Intelligence (ENSIA), \\
  Sidi Abdellah Campus,  Algiers, Algeria\\
  \texttt{aicha.boutorh@ensia.edu.dz} \\
  \And
 Mohamed Hadj Ameur \\
  National School of Artificial Intelligence (ENSIA), \\
  Sidi Abdellah Campus,  Algiers, Algeria\\
  \texttt{mohamed.hadj.ameur@ensia.edu.dz} \\
}

\begin{document}
\maketitle
\begin{abstract}

Large Language Models (LLMs) generate text autoregressively, relying on a key-value (KV) cache whose memory footprint grows linearly with context length, creating a major bottleneck. Recent compression methods mitigate this cost via token merging; however, these approaches often rely on indiscriminate aggregation, which degrades representations and introduces \emph{attention sag}, a mismatch where merged tokens receive the same softmax mass as individual tokens despite encoding multiple inputs. We propose a training-free, dual-component framework for KV cache compression that addresses these limitations. First, a \emph{soft cosine gate} adaptively modulates merging decisions based on value-vector similarity, suppressing or discarding dissimilar tokens to preserve semantic fidelity. Second, we introduce an \emph{attention-ratio} compensation mechanism that applies a decoding-time logit bias derived from prefill attention statistics, correcting the softmax imbalance induced by merging. Evaluated on LongBench (16 English datasets) while retaining only 25\% of the KV cache, our framework achieves strong compressed performance against representative one-shot baselines. It is especially robust on the evaluated grouped-query attention (GQA) models, maintaining near-lossless generation quality. Furthermore, the method outperforms the full-cache baseline on complex multi-document QA tasks and delivers a 3.3x decoding speedup at 100k tokens.

\end{abstract}



\section{Introduction}
\label{sec:intro}

Large language models (LLMs) achieve strong performance across diverse tasks, but their autoregressive inference is bottlenecked by the key-value (KV) cache, which stores past key and value vectors for attention computation. For a model with $L$ layers and $H$ attention heads of dimension $d$, the KV cache grows as $O(L \cdot H \cdot d \cdot n)$ with sequence length $n$, consuming multiple gigabytes even at moderate context lengths. As applications demand longer contexts, KV cache memory has become a primary constraint on batch size, throughput, and deployment cost.

Two families of methods have emerged for KV cache compression. \emph{Eviction} methods~\cite{zhang2023h2o,miao2024snapkv,xiao2024streamingllm,cai2024pyramidkv} permanently discard low-importance tokens, reducing cache size but irreversibly losing the information they carry. \emph{Merging} methods such as KVMerger~\cite{yuan2024kvmerger}, WeightedKV~\cite{yuan2024weightedkv}, D2O~\cite{wan2024d2o}, and LOOK-M~\cite{wan2024lookm} instead consolidate evicted tokens into retained cache entries, which can preserve more semantic content when information is distributed across many tokens.

We focus on two limitations of current merging methods:

\paragraph{Problem 1: The uniform merge-or-drop decision.}
Most methods apply the same strategy to \emph{all} evicted tokens: either merge all of them (KVMerger, WeightedKV, LOOK-M) or drop all of them~\cite{zhang2023h2o,miao2024snapkv,xiao2024streamingllm}. D2O~\cite{wan2024d2o} partitions tokens into drop and merge sets based on attention patterns, but this is a binary decision that still merges every token assigned to the merge set at full weight. No existing method modulates \emph{how much} to merge on a continuous, per-token basis using representation similarity. When an evicted token's value vector is dissimilar to its merge target, the weighted average can corrupt the retained representation. Our experiments illustrate this trade-off: uniform merging improves multi-document QA by $+0.92$ points on HotpotQA but degrades few-shot classification by $-4.0$ points on TREC. EMS~\cite{zhang2024ems} similarly notes that \emph{``both over-merging and over-evicting yield sub-optimal performance,''} yet does not provide a per-token mechanism.

\paragraph{Problem 2: Attention sag.}
When multiple tokens are merged into a single cache entry, that entry still receives roughly the same softmax attention as an unmerged token, despite representing several originals. KeepKV~\cite{li2024keepkv} formalizes this as \emph{attention sag} (Theorem~3.2): a position representing $m$ merged tokens is systematically under-attended by a factor of $\sim$$m$. As a result, many existing merging methods likely underuse the information they preserve~\cite{yuan2024kvmerger,yuan2024weightedkv,wan2024d2o,liu2025chunkkv}.

We address both problems with SelKV (Selective KV Cache Merging), a training-free framework with two complementary mechanisms that can be integrated into existing merging pipelines:

\begin{enumerate}
    \item \textbf{Soft cosine gate} (Problem~1): For each evicted token, we compute the cosine similarity between its value vector and its merge target. The gate $g = \max(\mathrm{cos\_sim}, 0)$ then modulates merge intensity continuously: similar tokens merge fully ($g \approx 1$), orthogonal tokens are dropped ($g = 0$), and intermediate cases merge partially. No learned parameters or threshold tuning are required.

    \item \textbf{Attention compensation} (Problem~2): We add an attention-ratio logit bias during decoding, using prefill attention weights to estimate how much attention mass each merged position should receive. Unlike na\"ive count-based compensation ($\log(1 + M_i)$), this formulation is naturally calibrated.
\end{enumerate}

Adaptive per-token routing has precedent in vision transformers, where DiffRate~\cite{chen2023diffrate} showed that learned merge-or-prune decisions can outperform uniform strategies. Our soft cosine gate is a training-free analogue for KV caches.

We introduce SelKV (Selective KV Cache Merging), a training-free framework with a \emph{soft cosine gate} that adaptively modulates per-token merge intensity and \emph{attention-ratio compensation} that recalibrates softmax attention toward merged positions. We evaluate on LongBench~\cite{bai2024longbench} (16 English datasets) with three models spanning MHA and GQA architectures at a 25\% KV retention budget. SelKV is especially effective on \emph{grouped-query attention} (GQA), where it comes closest to full-cache performance. It also exceeds the full-cache baseline on several multi-document QA settings, suggesting that selective merging can act as an implicit attention filter. At 100k tokens, the compressed cache decodes $3.3\times$ faster than the full cache.

The remainder of this paper is organized as follows: Section~\ref{sec:related_work} reviews related work, Section~\ref{sec:method} presents the proposed methodology; Section~\ref{sec:experiments} reports experimental evaluations and benchmarks against SOTA models; and Section~\ref{sec:conclusion} concludes with future directions .

\section{Related Work}
\label{sec:related_work}

KV cache compression methods largely fall into three groups: token eviction, token merging, and quantization. Quantization methods such as KIVI~\cite{liu2024kivi} and ZipCache~\cite{he2025zipcache} reduce precision rather than token count and are largely complementary to our method, so we focus here on eviction and merging.

\subsection{Token Eviction}
\label{sec:rw_eviction}

Token eviction methods reduce the KV cache by permanently discarding entries deemed unimportant, retaining only a subset for subsequent decoding.

Per-step eviction: H2O~\cite{zhang2023h2o} formulates cache eviction as a dynamic submodular problem, identifying \emph{Heavy Hitter} tokens that accumulate disproportionate attention mass and evicting the rest at every decoding step while preserving a recent-token window. VATP~\cite{guo2024vatp} extends this idea by observing that attention weight alone is an incomplete importance proxy: a token's actual contribution to the output is the product of its attention weight and its value magnitude. While effective, per-step methods re-evaluate importance and reorganize the cache at \emph{every} generation step, incurring substantial decode-time overhead.

One-shot eviction: An alternative family compresses the cache once after prefill, then generates with a fixed compressed cache. StreamingLLM~\cite{xiao2024streamingllm} keeps only the first few \emph{attention sink} tokens and a sliding recent window, requiring no attention-based scoring but sacrificing all middle-context information. SnapKV~\cite{miao2024snapkv} uses an observation window over the last $w$ query positions to identify important KV positions per head, smoothed by an average-pooling kernel to preserve token clusters. PyramidKV~\cite{cai2024pyramidkv} applies a layer-adaptive budget: lower layers retain more tokens, while upper layers rely more on recent context. More recent one-shot methods include RocketKV~\cite{behnam2025rocketkv}, EvolKV~\cite{cai2024evolkv}, and LAVa~\cite{shen2025lava}.

All eviction methods permanently destroy the information carried by discarded tokens. This can be especially costly when important evidence is spread across many positions, as in multi-document QA or long-range summarization.

\subsection{Token Merging}
\label{sec:rw_merging}

Rather than discarding evicted tokens entirely, merging methods consolidate their information into retained cache entries.

KVMerger~\cite{yuan2024kvmerger} identifies sequences of tokens with similar key representations and fuses them via a Gaussian kernel-weighted merging scheme. D2O~\cite{wan2024d2o} partitions tokens into those that should be dropped and those that should be merged based on attention patterns, but applies the \emph{same} strategy to all tokens within each partition. LOOK-M~\cite{wan2024lookm} merges tokens within a local sliding window to preserve spatial locality. WeightedKV~\cite{yuan2024weightedkv} applies attention-weighted convex combinations during merging. ChunkKV~\cite{liu2025chunkkv} treats fixed-size semantic chunks as compression units rather than individual tokens.

Most existing methods apply a \textbf{uniform} merge-or-drop decision: either \emph{all} evicted tokens are merged at full weight (KVMerger, WeightedKV, LOOK-M) or \emph{all} are dropped (H2O, SnapKV, StreamingLLM). D2O partitions tokens into drop and merge groups using attention patterns, but tokens assigned to the merge group are still merged unconditionally. EMS~\cite{zhang2024ems} explicitly observes that \emph{``both over-merging and over-evicting yield sub-optimal performance,''} yet proposes no continuous per-token gating mechanism. Our soft cosine gate fills this gap by modulating merge intensity on a $[0, 1]$ spectrum based on value-vector similarity.

A further problem affects all merging methods. When multiple tokens are consolidated into a single cache entry, the merged entry receives roughly the same softmax attention as a single unmerged token. KeepKV~\cite{li2024keepkv} formalizes this as \emph{attention sag} (Theorem~3.2): merged positions are systematically under-attended by a factor of $\sim$$m$. KeepKV proposes a logit bias of $\log(1 + M_i)$ based on raw merge counts as a correction. However, raw merge counts can be extreme (up to $\sim$200 at 75\% compression), making count-based compensation unstable. Moreover, the majority of merging methods like KVMerger, WeightedKV, D2O, ChunkKV do not incorporate any compensation at all.

\section{Method}
\label{sec:method}

We present \textbf{SelKV}, a pluggable framework for selective KV cache compression that addresses the two problems identified in Section~\ref{sec:intro}. Given a pretrained transformer with $L$ layers, $H$ attention heads, and head dimension $d$, SelKV compresses the KV cache \emph{once} after prefill and then decodes with the compressed cache. The framework introduces two mechanisms: (1)~a \emph{soft cosine gate} that decides per token how much to merge, and (2)~\emph{attention compensation} that corrects the under-attention of merged positions. Both are training-free and can be added to existing merging methods.

\begin{figure}[H]
  \centering
  \includegraphics[width=1\columnwidth]{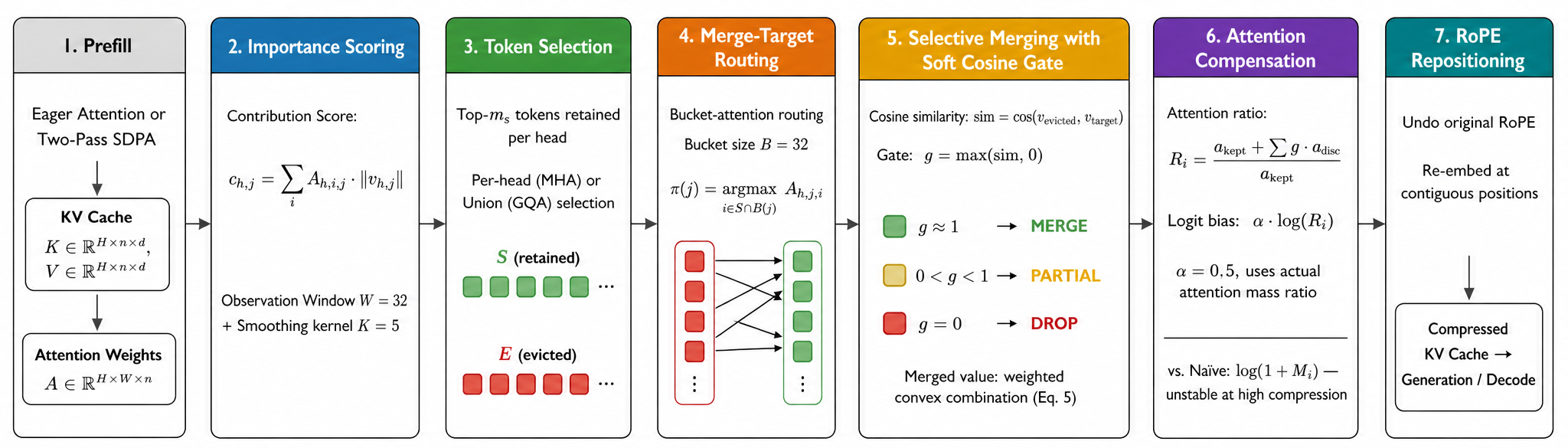}
  \caption{Overview of the selective KV cache compression pipeline. After prefill, the method scores token importance, selects retained tokens, routes evicted tokens to merge targets, applies the soft cosine gate, compensates attention for merged positions, and repositions RoPE indices before decode.}
  \label{fig:pipeline}
\end{figure}

Figure~\ref{fig:pipeline} illustrates the full pipeline, which proceeds in six stages after prefill: importance scoring (\S\ref{sec:scoring}), token selection (\S\ref{sec:selection}), merge-target routing (\S\ref{sec:routing}), selective merging with soft cosine gate (\S\ref{sec:soft_gate}), attention compensation (\S\ref{sec:compensation}), and RoPE repositioning (\S\ref{sec:rope}). Token selection supports both \emph{per-head} mode (each attention head independently selects its own tokens) and \emph{shared} mode (all heads retain the same token set), and attention compensation can be applied either averaged across layers or \emph{per-layer} via forward hooks.

\subsection{Overview and Notation}
\label{sec:overview}

Let $\mathbf{x} = (x_1, \dots, x_n)$ denote the input sequence of length $n$. We need (i)~the full KV cache $\{(\mathbf{K}^{(\ell)}, \mathbf{V}^{(\ell)})\}_{\ell=1}^{L}$, where $\mathbf{K}^{(\ell)}, \mathbf{V}^{(\ell)} \in \mathbb{R}^{H \times n \times d}$; and (ii)~attention weights over an observation window of $W$ query positions for importance scoring and routing. A na\"ive eager prefill would materialize $O(n^2)$ attention, which is infeasible at long contexts. Instead, we use a \emph{two-pass} strategy: (1)~run SDPA prefill to build the full KV cache without materializing attention matrices, then (2)~re-run only the last $W$ tokens with eager attention against the cached keys, producing attention matrices of size $H \times W \times n$ per layer. This reduces prefill memory from $O(n^2)$ to $O(W \cdot n)$ while yielding the same attention weights used by SnapKV-style observation windows.

Given a target retention ratio $\rho \in (0, 1)$, we set the target cache length to $m = \lfloor \rho \cdot n \rfloor$. A \emph{recent window} of the last $m_r = 16$ tokens is always retained to preserve local context~\cite{xiao2024streamingllm}. The remaining $m_s = m - m_r$ slots are filled by importance-based selection.

After compression, we switch to SDPA for decode, which does not require materializing full attention matrices and benefits from the reduced cache size.

\subsection{Importance Scoring}
\label{sec:scoring}

Prior methods score token importance by cumulative attention alone~\cite{zhang2023h2o}. Following the insight of VATP~\cite{guo2024vatp}, we observe that attention weight alone is an incomplete proxy: a token may receive high attention but carry a near-zero value vector, contributing little to the output. Conversely, a moderately attended token with a large value vector may have substantial impact.

We define the \emph{contribution score} of token $j$ at layer $\ell$ and head $h$ as:
\begin{equation}
\label{eq:contribution}
c_{h,j}^{(\ell)} = \sum_{i \in \mathcal{W}} A_{h,i,j}^{(\ell)} \cdot \| \mathbf{v}_{h,j}^{(\ell)} \|_2
\end{equation}
where $\mathcal{W}$ denotes the set of query positions used for scoring. Following SnapKV~\cite{miao2024snapkv}, we restrict $\mathcal{W}$ to the last $W = 32$ positions and apply average-pooling smoothing with kernel size $K = 5$ to preserve token clusters.

\subsection{Token Selection}
\label{sec:selection}

Different layers exhibit distinct attention patterns: lower layers tend to have broader, more diffuse attention, while upper layers focus more narrowly~\cite{cai2024pyramidkv}. We therefore perform token selection independently at each layer.

\paragraph{Per-head selection (MHA models).}
Each head $h$ at layer $\ell$ independently selects its top-$m_s$ tokens by contribution score:
\begin{equation}
\label{eq:perhead_selection}
\mathcal{S}_h^{(\ell)} = \operatorname{top}\nolimits_{m_s}\!\bigl( \{c_{h,j}^{(\ell)}\}_{j=1}^{n - m_r} \bigr) \;\cup\; \{n - m_r + 1, \dots, n\}
\end{equation}
The complementary set $\mathcal{E}_h^{(\ell)} = \{1, \dots, n\} \setminus \mathcal{S}_h^{(\ell)}$ contains the evicted tokens for head $h$.

\paragraph{Union selection (GQA models).}
On GQA models, each KV head serves $G$ query heads. Each query head produces its own importance scores $c_{g,j}$; we first average these across the $G$ query heads sharing a KV head to obtain per-KV-head scores. Each KV head then independently selects its top-$m_s$ tokens, and we take the \emph{set union} across heads: a token is kept if \emph{any} KV head selects it. If the union set exceeds the layer budget $m$, we trim by the mean aggregate score across heads until the budget is met. This ensures that tokens critical to even a single head are retained, at the cost of a slightly variable per-layer cache size before trimming.

\subsection{Merge-Target Routing}
\label{sec:routing}

Each evicted token $j \in \mathcal{E}_h^{(\ell)}$ must be assigned a merge target $\pi_h(j) \in \mathcal{S}_h^{(\ell)}$. We use \emph{bucket-attention routing}: the sequence is partitioned into positional buckets of size $B = 32$, and each evicted token is routed to the kept token within the same bucket that received the highest attention from it during prefill:
\begin{equation}
\label{eq:routing}
\pi_h(j) = \operatorname*{argmax}_{i \in \mathcal{S}_h^{(\ell)} \cap \mathcal{B}(j)} A_{h,j,i}^{(\ell)}
\end{equation}
This uses the $H \times W \times n$ attention matrix materialized during the observation-window pass (Section~\ref{sec:overview}): for each evicted token $j$, we look up the attention that the observation-window queries paid to $j$ and to the candidate targets within its bucket. The complexity is $O(H \cdot |\mathcal{E}| \cdot B)$, reduced from $O(H \cdot |\mathcal{S}| \cdot |\mathcal{E}| \cdot d)$ for full cosine routing.

\subsection{Selective Merging with Soft Cosine Gate}
\label{sec:soft_gate}

Uniformly merging all evicted tokens is harmful when an evicted token's value vector is dissimilar to its merge target's value vector. We need a mechanism that merges when similarity is high and drops when similarity is low, without any learned parameters.

For each evicted token $j$ routed to target $\pi_h(j)$ at head $h$, we compute the cosine similarity between their value vectors:
\begin{equation}
\label{eq:cosine_sim}
\mathrm{sim}_{h,j} = \frac{\mathbf{v}_{h,j}^{(\ell)} \cdot \mathbf{v}_{h,\pi_h(j)}^{(\ell)}}{\| \mathbf{v}_{h,j}^{(\ell)} \|_2 \cdot \| \mathbf{v}_{h,\pi_h(j)}^{(\ell)} \|_2}
\end{equation}
The soft gate is $g_{h,j} = \max(\mathrm{sim}_{h,j},\; 0)$. It operates on a continuous spectrum: $g = 0$ (drop) when vectors are dissimilar, $0 < g < 1$ (partial merge) for moderate similarity, and $g \approx 1$ (full merge) for high similarity.

The merged value for a kept token $i$ is:
\begin{equation}
\label{eq:merge}
\tilde{\mathbf{v}}_{h,i}^{(\ell)} = \frac{a_{h,i} \cdot \mathbf{v}_{h,i} \;+\; \sum_{j : \pi_h(j) = i} g_{h,j} \cdot a_{h,j} \cdot \mathbf{v}_{h,j}}{a_{h,i} \;+\; \sum_{j : \pi_h(j) = i} g_{h,j} \cdot a_{h,j}}
\end{equation}
where $a_{h,i} = \sum_{q \in \mathcal{W}} A_{h,q,i}$ is the cumulative attention over the observation window. When $g = 0$ for all evicted tokens, Eq.~\ref{eq:merge} reduces to pure drop; when $g = 1$, it reduces to standard attention-weighted merging. The gate is training-free, parameter-free, and operates per token and per head.

\subsection{Attention Compensation}
\label{sec:compensation}

As established by KeepKV~\cite{li2024keepkv} (Theorem~3.2), softmax attention systematically under-weights merged positions. KeepKV compensates by adding $\log(1 + M_i)$ to attention logits, where $M_i$ is the raw merge count. Under heavy compression ($\ge$75\% eviction), however, merge counts can reach $\sim$200, so count-based correction becomes less stable. We therefore replace raw counts with an \emph{attention-ratio} derived from the prefill pass:
\begin{equation}
\label{eq:compensation}
R_{h,i} = \frac{a_{h,i}^{\mathrm{kept}} + \sum_{j \to i} g_{h,j} \cdot a_{h,j}^{\mathrm{disc}}}{a_{h,i}^{\mathrm{kept}}}
\end{equation}
On GQA models (where all query heads sharing a KV head operate on the same retained token set), we average $R_{h,i}$ across KV heads and add $\alpha \cdot \log(R_i)$ (with $\alpha = 0.5$) to the attention logits during decoding. On MHA models with per-head selection, the bias is applied per head without cross-head averaging, since each head retains a different token set. Unlike raw merge counts, $R_i$ stays bounded by the redistributed attention mass, so positions that absorbed negligible attention receive little or no boost.

\subsection{RoPE Repositioning}
\label{sec:rope}

After compression, we re-embed retained keys at contiguous positions by undoing the original RoPE and applying new position embeddings:
\begin{equation}
\label{eq:rope}
\tilde{\mathbf{k}}_i = \mathrm{RoPE}\!\bigl(\mathrm{RoPE}^{-1}(\mathbf{k}_i,\, p_i^{\mathrm{old}}),\; p_i^{\mathrm{new}}\bigr)
\end{equation}
In our ablation, disabling repositioning causes repetition loops in approximately 15\% of generated outputs on short-context models. However, on long-context models under heavy compression ($>$75\% eviction), repositioning can distort positional identity and harm retrieval tasks (Section~\ref{sec:ablation}). We therefore disable it by default for long-context evaluation.

\subsection{Algorithm Summary}

The full pipeline is presented in Algorithm~\ref{alg:pipeline}.

\begin{algorithm}[H]
\small
\fbox{\parbox{0.95\columnwidth}{
\textbf{Selective KV Cache Compression}\\[2pt]
\textbf{Input:} KV cache $\{(\mathbf{K}^{(\ell)}, \mathbf{V}^{(\ell)})\}$, attention $\{\mathbf{A}^{(\ell)}\}$, ratio $\rho$\\
\textbf{Output:} Compressed KV cache, attention bias vector\\[2pt]
\textbf{for} each layer $\ell = 1, \dots, L$ \textbf{do}\\
\hspace*{1em}1. Score tokens: $c_{h,j}^{(\ell)} = \sum_{i \in \mathcal{W}} A_{h,i,j} \cdot \|\mathbf{v}_{h,j}\|$ \hfill \textit{(Eq.~\ref{eq:contribution})}\\
\hspace*{1em}2. Select top-$m_s$ per head: $\mathcal{S}_h^{(\ell)}$; evict $\mathcal{E}_h^{(\ell)}$ \hfill \textit{(Eq.~\ref{eq:perhead_selection})}\\
\hspace*{1em}\textbf{for} each evicted $j \in \mathcal{E}_h^{(\ell)}$ \textbf{do}\\
\hspace*{2em}3. Route to target $\pi_h(j)$ via bucket-attention \hfill \textit{(Eq.~\ref{eq:routing})}\\
\hspace*{2em}4. Gate: $g_{h,j} = \max(\mathrm{sim}_{h,j},\, 0)$ \hfill \textit{(Eq.~\ref{eq:cosine_sim})}\\
\hspace*{1em}\textbf{end for}\\
\hspace*{1em}5. Merge: $\tilde{\mathbf{v}}_{h,i} \leftarrow$ gated attn-weighted avg \hfill \textit{(Eq.~\ref{eq:merge})}\\
\hspace*{1em}6. Compute attn ratio $R_{h,i}$ per kept position \hfill \textit{(Eq.~\ref{eq:compensation})}\\
\hspace*{1em}7. Reposition keys via RoPE inverse + re-embed \hfill \textit{(Eq.~\ref{eq:rope})}\\
\textbf{end for}\\
\textbf{Decode} with compressed cache + $\alpha \cdot \log(R_i)$ bias
}}
\caption{Pseudocode for selective KV cache compression.}
\label{alg:pipeline}
\end{algorithm}

\section{Experiments}
\label{sec:experiments}

\subsection{Setup}
\label{sec:setup}

\paragraph{Models.}
We evaluate on three models spanning two attention architectures: LongChat-7B-v1.5-32k (multi-head attention, MHA, 32 KV heads), LLaMA-3.1-8B-Instruct (grouped-query attention, GQA, 8 KV heads), and Gemma-2-9B-IT (GQA, 8 KV heads with alternating sliding-window attention). LongChat uses 31{,}500-token inputs, enabling evaluation under heavy compression (75\% eviction); the remaining models use 3{,}500-token inputs. All experiments use FP16 inference on NVIDIA H100 80GB GPUs.

\paragraph{Benchmark.}
We evaluate on the 16 English datasets of LongBench~\cite{bai2024longbench}, spanning 6 task categories: single-document QA (NarrativeQA, Qasper, MultifieldQA-en), multi-document QA (HotpotQA, 2WikiMQA, MuSiQue), summarization (GovReport, QMSum, MultiNews), few-shot learning (TREC, TriviaQA, SAMSum), synthetic tasks (PassageCount, PassageRetrieval-en), and code completion (LCC, RepoBench-P). Each dataset is evaluated with its standard metric (F1, ROUGE, accuracy, or code similarity) on 200 samples (500 for code tasks).

\paragraph{Baselines.}
We compare against three one-shot methods at a 25\% KV retention budget: SnapKV~\cite{miao2024snapkv} (observation-window eviction), LOOK-M~\cite{wan2024lookm} (sliding-window merging), and PyramidKV~\cite{cai2024pyramidkv} (layer-adaptive eviction). Section~\ref{sec:latency} reports decode throughput for SelKV against the full-cache baseline.

\paragraph{SelKV configuration.}
Contribution-based importance scoring (attention $\times$ value norm), observation window $W = 32$, smoothing kernel $K = 5$, bucket-attention routing with bucket size $B = 32$, soft cosine gate enabled, attention compensation ($\alpha{=}0.5$, attn-ratio), recent window $m_r = 16$ tokens, RoPE repositioning disabled. On GQA models, each KV head independently selects its top-$m_s$ tokens and the set union across heads determines the kept set (trimmed to budget by mean score if needed).

\subsection{Cross-Model Results}
\label{sec:cross_model}

We evaluate our approach SelKV and three one-shot baselines across three LLMs---Gemma-2-9B-IT, LLaMA-3.1-8B-Instruct, and LongChat-7B-v1.5-32k---covering six LongBench task categories: single-document QA, multi-document QA, summarization, few-shot learning, synthetic tasks, and code completion. Table~\ref{tab:full_results} reports the per-dataset results for all methods and models, while the final Avg, $\Delta$, and Avg Rank columns summarize overall performance and consistency within each model. Readers should focus on whether SelKV stays closest to the Full Cache baseline across architectures and where compression is most beneficial, especially on GQA models and multi-document QA tasks.

\begin{table}[H]
\centering
\caption{LongBench results (16 English datasets) at a 25\% KV retention budget across three models. $\Delta$: gap to Full Cache within each model. Avg Rank summarizes per-dataset ranks among compressed methods within each model; it is the mean rank across datasets (lower is better), and ties receive the better rank. Best compression method per task in \textbf{bold}.}
\label{tab:full_results}
\scriptsize
\setlength{\tabcolsep}{2.5pt}
\resizebox{\textwidth}{!}{%
\begin{tabular}{ll|ccc|ccc|ccc|ccc|cc|cc|ccc}
\toprule
& & \multicolumn{3}{c|}{Single-Doc QA} & \multicolumn{3}{c|}{Multi-Doc QA} & \multicolumn{3}{c|}{Summarization} & \multicolumn{3}{c|}{Few-shot} & \multicolumn{2}{c|}{Synthetic} & \multicolumn{2}{c|}{Code} & \multicolumn{3}{c}{Overall} \\
\textbf{LLMs} & \textbf{Method} & \rotatebox{60}{NrtQA} & \rotatebox{60}{Qasp} & \rotatebox{60}{MF-en} & \rotatebox{60}{HpQA} & \rotatebox{60}{2Wiki} & \rotatebox{60}{Musi} & \rotatebox{60}{GovR} & \rotatebox{60}{QMSm} & \rotatebox{60}{MNws} & \rotatebox{60}{TREC} & \rotatebox{60}{TrvQA} & \rotatebox{60}{SAMs} & \rotatebox{60}{PCnt} & \rotatebox{60}{PRet} & \rotatebox{60}{LCC} & \rotatebox{60}{RB-P} & \textbf{Avg} & $\Delta$ & \textbf{Avg Rank} \\
\midrule
\multirow{5}{*}{\rotatebox{90}{Gemma-2}} & Full Cache & 22.93 & 40.79 & 46.68 & 47.25 & 48.74 & 22.84 & 27.63 & 19.47 & 24.81 & 67.50 & 91.92 & 43.56 & 7.00 & 33.50 & 72.78 & 70.33 & 42.98 & -- & -- \\
 & SnapKV & 22.22 & 40.01 & \textbf{46.78} & 46.90 & 46.54 & 22.00 & 24.63 & 19.28 & 22.30 & 66.50 & 91.92 & 42.62 & 7.00 & 33.50 & \textbf{72.62} & 69.75 & 42.16 & $-$0.82 & 2.06 \\
 & LOOK-M & 22.66 & 36.75 & 45.93 & \textbf{46.97} & 46.05 & 22.34 & 23.98 & 19.18 & 22.73 & 55.50 & \textbf{92.17} & 42.25 & 7.00 & 33.50 & 72.07 & \textbf{69.84} & 41.18 & $-$1.80 & 2.56 \\
 & PyramidKV & 22.43 & 39.70 & 46.71 & 46.37 & \textbf{47.30} & 22.41 & 24.46 & 18.94 & 22.12 & 66.50 & 91.92 & 42.36 & 7.00 & 33.50 & 72.53 & 69.10 & 42.08 & $-$0.90 & 2.50 \\
 & \textbf{SelKV(ours)} & \textbf{23.32} & \textbf{40.01} & 46.44 & 46.67 & 46.84 & \textbf{22.90} & \textbf{24.94} & \textbf{19.33} & \textbf{22.79} & \textbf{67.00} & 91.76 & \textbf{42.89} & \textbf{7.00} & \textbf{33.50} & 72.24 & 69.35 & \textbf{42.31} & \textbf{$-$0.67} & \textbf{1.69} \\
\midrule
\multirow{5}{*}{\rotatebox{90}{\scriptsize LLaMA-3.1}} & Full Cache & 21.63 & 40.24 & 49.54 & 39.04 & 35.23 & 18.84 & 32.37 & 21.00 & 27.15 & 70.00 & 89.00 & 43.68 & 3.58 & 34.00 & 64.87 & 58.41 & 40.54 & -- & -- \\
 & SnapKV & 21.33 & \textbf{39.00} & 49.71 & 39.25 & 35.86 & 18.88 & 27.97 & 21.04 & 23.59 & \textbf{67.50} & 89.63 & 42.00 & \textbf{4.08} & 34.00 & \textbf{65.02} & \textbf{57.73} & 39.79 & $-$0.75 & 2.06 \\
 & LOOK-M & 20.84 & 33.18 & 44.92 & 38.96 & 35.36 & 17.01 & \textbf{28.33} & \textbf{21.59} & \textbf{24.84} & 65.00 & 89.35 & 41.97 & 4.00 & \textbf{35.00} & 63.80 & 56.31 & 38.78 & $-$1.76 & 3.00 \\
 & PyramidKV & 21.60 & 38.04 & 49.20 & \textbf{40.04} & 34.77 & 18.65 & 27.63 & 20.76 & 23.34 & 67.00 & 89.77 & \textbf{42.37} & \textbf{4.08} & 34.00 & 63.83 & 57.19 & 39.52 & $-$1.02 & 2.63 \\
 & \textbf{SelKV(ours)} & \textbf{21.60} & 38.25 & \textbf{50.01} & 39.44 & \textbf{35.91} & \textbf{19.55} & 27.94 & 21.37 & 23.63 & 67.00 & \textbf{89.95} & 42.06 & 4.03 & 34.50 & 64.34 & 57.31 & \textbf{39.81} & \textbf{$-$0.73} & \textbf{1.75} \\
\midrule
\multirow{5}{*}{\rotatebox{90}{LongChat}} & Full Cache & 21.01 & 29.31 & 43.37 & 33.08 & 23.44 & 14.15 & 30.91 & 22.67 & 26.60 & 66.50 & 84.09 & 22.16 & 0.00 & 30.50 & 50.06 & 58.17 & 34.75 & -- & -- \\
 & SnapKV & \textbf{20.71} & 27.33 & 39.42 & \textbf{34.73} & 22.66 & \textbf{14.18} & 26.52 & 22.54 & 13.25 & 64.00 & \textbf{83.78} & 27.68 & 0.00 & 29.00 & 43.35 & \textbf{57.96} & 32.94 & $-$1.81 & 2.00 \\
 & LOOK-M & 19.86 & 23.16 & 29.05 & 31.78 & 21.58 & 11.93 & 22.55 & 22.39 & 15.24 & 55.50 & 74.33 & 25.93 & 0.00 & 23.25 & 47.76 & 56.86 & 30.07 & $-$4.68 & 3.38 \\
 & PyramidKV & 20.17 & \textbf{28.00} & \textbf{41.68} & 34.04 & \textbf{22.76} & 13.97 & \textbf{27.29} & \textbf{23.07} & \textbf{22.60} & \textbf{64.50} & 82.34 & 23.79 & 0.00 & 30.50 & \textbf{50.87} & 57.55 & \textbf{33.95} & \textbf{$-$0.80} & \textbf{1.63} \\
 & \textbf{SelKV(ours)} & 20.48 & 27.77 & 33.69 & 31.02 & 20.96 & 13.11 & 25.66 & 22.24 & 18.66 & 61.00 & 78.04 & \textbf{29.54} & \textbf{0.00} & \textbf{38.50} & 48.50 & 56.18 & 32.83 & $-$1.92 & 2.63 \\
\bottomrule
\end{tabular}
}
\end{table}

\paragraph{Architecture dependence.}
SelKV gives the top compressed result on both GQA models: Gemma-2 (42.31, $-$0.67) and LLaMA-3.1 (39.81, $-$0.73). On the MHA model LongChat (evaluated at 31{,}500 tokens vs.\ 3{,}500 for the GQA models), PyramidKV leads (33.95 vs.\ 32.83). To control for context length, we also evaluate on LLaMA-3.1 at 31{,}500 tokens (Appendix~\ref{app:longcontext}): SelKV (48.46) and SnapKV (48.46) are tied, both ahead of PyramidKV (48.26) and LOOK-M (47.16), showing that the advantage on GQA models is not an artifact of the shorter evaluation context.

\paragraph{Multi-document QA.}
On several multi-document QA settings, the compressed cache exceeds full-cache scores (e.g., LLaMA-3.1: HotpotQA 39.44 vs.\ 39.04, MuSiQue 19.55 vs.\ 18.84; Gemma-2: MuSiQue 22.90 vs.\ 22.84). We interpret this as evidence that our approach can suppress distracting context while preserving salient evidence.

\begin{figure}[H]
\centering
\includegraphics[width=\columnwidth]{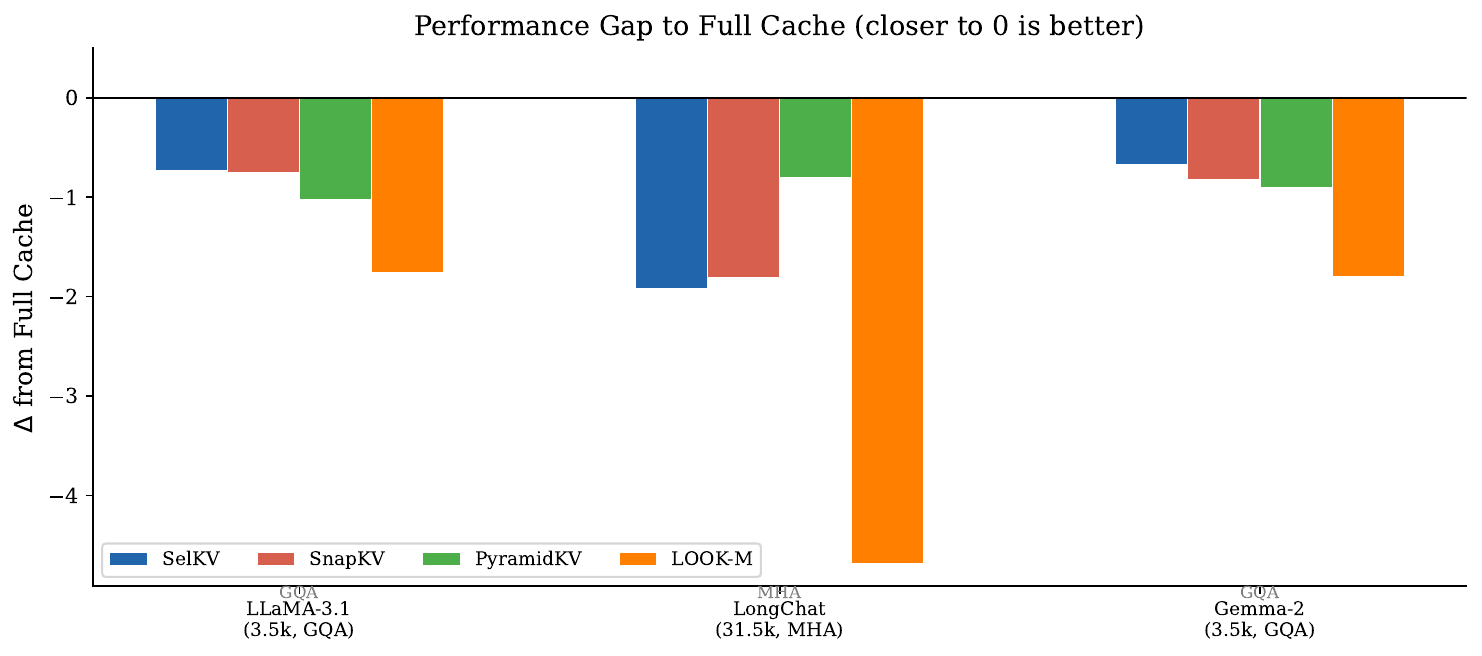}
\caption{Performance gap to Full Cache across models (closer to 0 is better). SelKV(ours) is the best for LLaMa and Gemma}
\label{fig:delta_bars}
\end{figure}


At the category level on Gemma-2, our approach SelKV leads on four of six categories (Single-Doc QA, Multi-Doc QA, Summarization, and Few-shot) while SnapKV leads only on Code, where exact token identity matters (full breakdown in Appendix~\ref{app:category}, Table~\ref{tab:category}).

\subsection{Ablation and Compression Ratio Robustness}
\label{sec:ablation}

We perform a staged ablation on LLaMA-3.1-8B (GQA, 4:1 ratio) at a 25\% KV retention budget over all 16 datasets. Starting from our token selection, we compare: (i)~\emph{eviction only}, which drops all evicted tokens; (ii)~\emph{merge all}, which merges without the soft cosine gate or attention compensation; (iii)~\emph{+ gate}, which adds the soft cosine gate; and (iv)~\emph{+ gate + comp}, which recovers the full method by further adding attention compensation. To test portability, we also evaluate the gate on a 6-dataset Gemma-2 subset and as a plug-in modification to LOOK-M. Finally, we test robustness to the retention budget by sweeping 10\%--90\% KV retention on LLaMA-3.1 and Gemma-2, with full per-ratio tables in Appendix~\ref{app:ratio_tables}.

Blind merging hurts on LLaMA-3.1 (39.01 vs.\ 39.69 for eviction only), consistent with frequent value mismatch under 4:1 GQA. The soft cosine gate recovers most of the loss (39.57, $+$0.56), and adding compensation yields further improvement (39.81, $+$0.24). On a 6-dataset subset of Gemma-2, compensation helps more than the gate ($+$0.08 vs.\ $-$0.02), suggesting that the two components matter differently across GQA ratios. The gate is also portable: adding it to LOOK-M improves Gemma-2 from 41.18 to 41.35 and LLaMA-3.1 from 38.78 to 38.86.

\begin{table}[H]
\centering
\caption{Ablation on LLaMA-3.1-8B (16 datasets, 25\% KV retention budget). ``Evict only'' uses our token selection but drops evicted tokens without merging.}
\label{tab:ablation_llama31}
\small
\begin{tabular}{lcccr}
\toprule
\textbf{Configuration} & \textbf{Gate} & \textbf{Comp} & \textbf{Merge} & \textbf{Avg (16 ds)} \\
\midrule
Full Cache              & --          & --          & --    & 40.54 \\
\midrule
Eviction only           & --          & --          & --    & 39.69 \\
Merge all (no gate)     & --          & --          & \checkmark & 39.01 \small{($-$0.68)} \\
\;\;+ gate              & \checkmark   & --          & \checkmark & 39.57 \small{(+0.56)} \\
\;\;+ gate + comp       & \checkmark   & \checkmark   & \checkmark & \textbf{39.81} \small{(+0.24)} \\
\bottomrule
\end{tabular}
\end{table}

\paragraph{RoPE repositioning.}
On LongChat-7B (31.5k tokens), repositioning harms PassageRetrieval-en (5.0 vs.\ 38.5 without) where paragraph identity depends on absolute position, so we disable it by default.

\paragraph{Compression-ratio robustness.}
We sweep the KV retention budget from 10\% to 90\% on LLaMA-3.1 and Gemma-2 (Figure~\ref{fig:ratio_sweep}, Appendix~\ref{app:ratio_tables}) to test whether the relative ranking is stable across compression levels. At 10\%, SnapKV leads on both models because few merge targets remain. From 25\% onward, SelKV leads on Gemma-2 and matches SnapKV on LLaMA-3.1; above 70\%, all methods converge. Notably, SelKV at an 80\% retention budget on Gemma-2 slightly exceeds the full cache (43.01 vs.\ 42.98).

\subsection{Latency Analysis}
\label{sec:latency}

We next evaluate whether one-shot compression translates into practical wall-clock gains as prompt length grows.

\begin{figure}[H]
\centering
\includegraphics[width=0.82\linewidth]{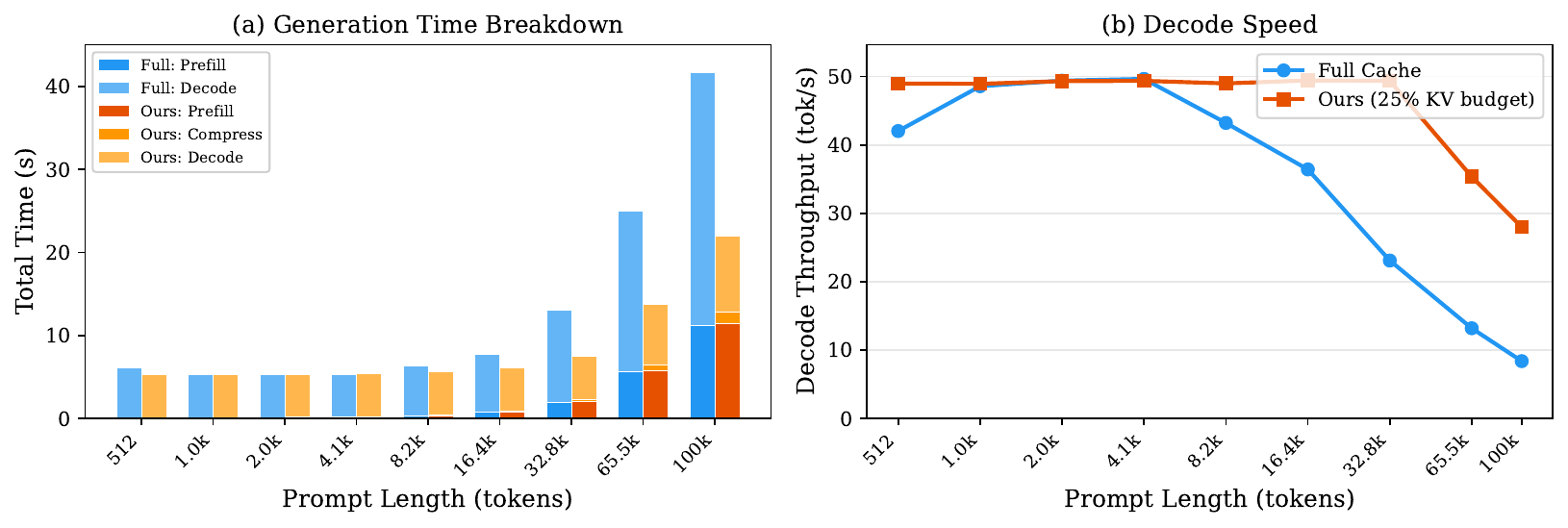}
\vspace{-3mm}
\caption{Latency on LLaMA-3.1-8B (256 generated tokens, H100, 25\% KV retention budget). (a)~Total time breakdown. (b)~Decode throughput.}
\label{fig:latency}
\end{figure}

Figure~\ref{fig:latency} shows wall-clock time on LLaMA-3.1-8B across prompt lengths from 512 to 100k tokens. At short contexts ($\le$8k), throughput matches the full cache ($\sim$50\,tok/s). Beyond 16k tokens, the smaller KV cache makes decoding faster: at 100k tokens we observe 28\,tok/s vs.\ 8.4\,tok/s for the full cache ($3.3\times$). Compression overhead remains below 5\% of total time at 8k+.

\section{Conclusion}
\label{sec:conclusion}

We present SelKV, a training-free KV-cache compression framework with a soft cosine gate and attention-ratio compensation. On LongBench at a 25\% KV retention budget, it gives the best compressed results on the two GQA models we study, stays within 0.67 - 0.73 points of full cache on average, slightly exceeds full-cache scores on some multi-document QA settings, and yields $3.3\times$ faster decoding at 100k tokens. These results suggest that merge quality, not just token selection, is an important design axis for KV compression. The cosine gate spans the continuum between pure eviction ($g{=}0$) and full merging ($g{=}1$), and offers a simple interface for future learned or task-adaptive policies.

Limitations include reduced effectiveness on MHA models and code tasks, and at very aggressive compression (10\% budget) where few merge targets remain. We also evaluate only three models, and generation quality is validated on contexts up to 31{,}500 tokens; the 100k-token latency benchmark measures throughput scaling only. Future work includes broader model coverage, combinations with quantization, and extensions to larger-context training-time adaptations and longer-context quality evaluation.


\bibliographystyle{plainnat}
\bibliography{references}

\appendix

\section{Computational Complexity}
\label{app:complexity}

The dominant prefill cost is the SDPA forward pass, $O(L \cdot H \cdot n^2 \cdot d)$, identical to any method that builds a KV cache. The two-pass eager re-pass over the observation window adds $O(L \cdot H \cdot W \cdot n \cdot d)$, which is linear in $n$ for fixed $W$. The compression step itself is efficient:

\begin{itemize}
    \item \textbf{Importance scoring}: $O(L \cdot H \cdot n \cdot (W + d))$  attention summation over the observation window and value-norm computation.
    \item \textbf{Selection}: $O(L \cdot H_{\mathrm{kv}} \cdot n \log m)$  per-KV-head top-$k$ via partial sort, followed by set union.
    \item \textbf{Routing}: $O(L \cdot H \cdot |\mathcal{E}| \cdot B)$ where $B$ is the bucket size (default 32).
    \item \textbf{Soft gate}: $O(L \cdot H \cdot |\mathcal{E}| \cdot d)$  one cosine similarity per evicted token.
    \item \textbf{Merge}: $O(L \cdot H \cdot |\mathcal{E}| \cdot d)$  vectorized scatter-add operations.
    \item \textbf{RoPE repositioning}: $O(L \cdot H \cdot m \cdot d)$.
    \item \textbf{Attention compensation} (decode): $O(m)$ per step  add a scalar bias per position.
\end{itemize}
The total compression overhead is 0.04\,s for a 2048-token sequence on an H100 GPU, amortized over the entire decode phase.

\section{Hyperparameter Sensitivity}
\label{app:hyperparams}

Table~\ref{tab:hyperparams} varies the observation window size $W$ and smoothing kernel size $K$ on Gemma-2 (6-dataset subset, 25\% KV retention budget).

\begin{table}[H]
\centering
\caption{Sensitivity to observation window $W$ and kernel size $K$ on Gemma-2-9B-IT (6 datasets, 25\% KV retention budget). Default: $W{=}32$, $K{=}5$.}
\label{tab:hyperparams}
\small
\begin{tabular}{cc|cccccc|c}
\toprule
$W$ & $K$ & Qasp & SAMs & HpQA & LCC & TREC & PRet & \textbf{Avg} \\
\midrule
16 & 5 & 40.16 & 43.04 & \textbf{47.42} & 72.29 & 67.00 & 33.00 & 50.49 \\
\textbf{32} & \textbf{5} & 40.01 & 42.89 & 46.67 & 72.24 & 67.00 & \textbf{33.50} & 50.39 \\
64 & 5 & 39.71 & \textbf{43.24} & \textbf{47.50} & 72.07 & 66.50 & 33.00 & 50.34 \\
\midrule
32 & 3 & \textbf{40.15} & 42.74 & 46.90 & 71.98 & \textbf{70.59} & \textbf{33.50} & \textbf{50.98} \\
32 & 7 & 39.83 & 42.73 & 46.50 & \textbf{72.50} & 67.50 & 33.00 & 50.34 \\
\bottomrule
\end{tabular}
\end{table}

All configurations fall within a 0.6-point range (50.34--50.98), demonstrating that our method is robust to these hyperparameter choices. The observation window has minimal effect ($W{=}16$ and $W{=}64$ perform comparably to $W{=}32$), and kernel sizes $K \in \{3, 5, 7\}$ yield near-identical results.

\section{Category-Level Analysis}
\label{app:category}

Table~\ref{tab:category} summarizes per-category averages on Gemma-2-9B-IT (25\% KV retention budget). Our method leads on four of six categories: Single-Doc QA, Multi-Doc QA, Summarization, and Few-shot. SnapKV leads only on Code, where precise token identity matters.

\begin{table}[H]
\centering
\caption{Category-level averages on LongBench (Gemma-2-9B-IT, 25\% KV retention budget). Best one-shot result per category in \textbf{bold}.}
\label{tab:category}
\small
\begin{tabular}{l|ccccc}
\toprule
\textbf{Category} & \textbf{Full} & \textbf{SelKV(ours)} & \textbf{SnapKV} & \textbf{LOOK-M} & \textbf{PyramidKV} \\
\midrule
Single-Doc QA   & 36.80 & \textbf{36.59} & 36.34 & 35.11 & 36.28 \\
Multi-Doc QA    & 39.61 & \textbf{38.80} & 38.48 & 38.45 & 38.69 \\
Summarization   & 23.97 & \textbf{22.35} & 22.07 & 21.96 & 21.84 \\
Few-shot        & 67.66 & \textbf{67.22} & 67.01 & 63.31 & 66.93 \\
Synthetic       & 20.25 & \textbf{20.25} & 20.25 & 20.25 & 20.25 \\
Code            & 71.56 & 70.80 & \textbf{71.19} & 70.96 & 70.82 \\
\bottomrule
\end{tabular}
\end{table}

\section{Compression Ratio Sweep}
\label{app:ratio_tables}
\begin{figure}[H]
\centering
\includegraphics[width=\columnwidth]{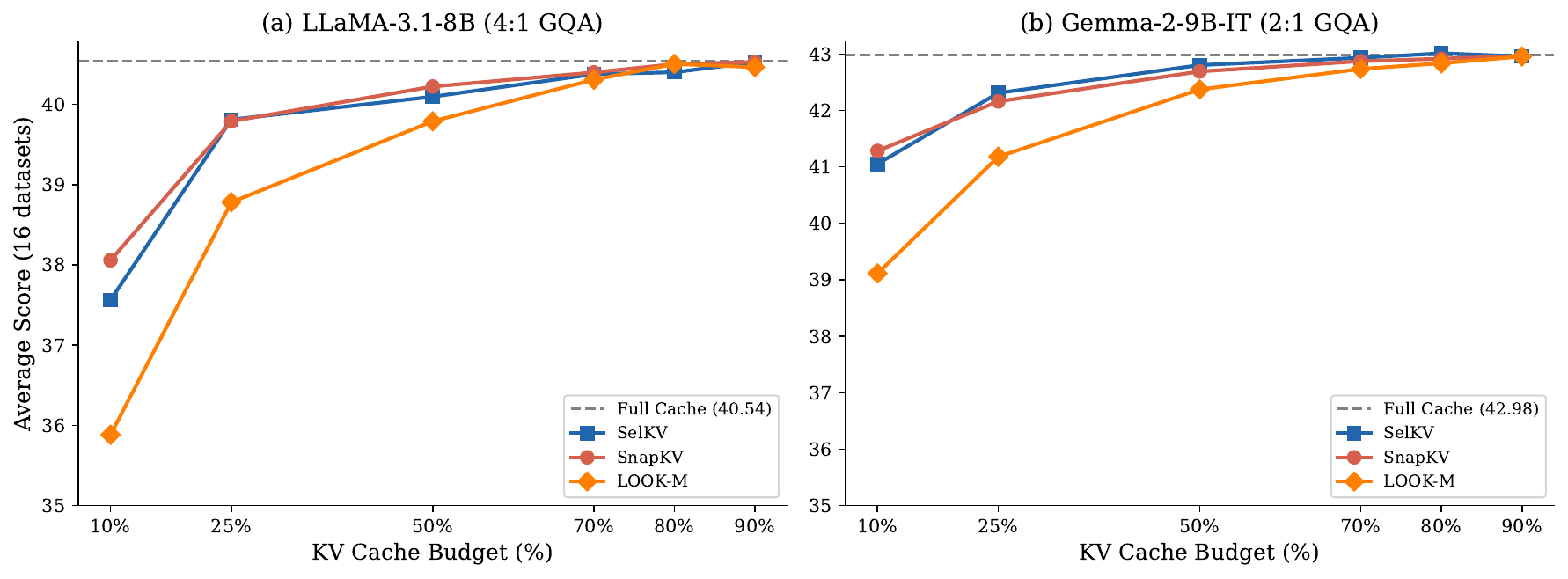}
\caption{LongBench average at varying KV retention ratios (3.5k tokens, 16 datasets). (a)~LLaMA-3.1-8B (b)~Gemma-2-9B-IT.}

\label{fig:ratio_sweep}
\end{figure}

Tables~\ref{tab:ratio_llama} and~\ref{tab:ratio_gemma} present per-method scores at six compression ratios on LLaMA-3.1-8B and Gemma-2-9B-IT (3.5k tokens, 16 datasets each). On LLaMA-3.1 (Figure~\ref{fig:ratio_sweep}), SnapKV leads at 10\%; from 25\% onward, SelKV matches or exceeds SnapKV; above 70\%, all methods converge. On Gemma-2, SelKV leads at every budget from 25\% onward, and at 80\% it slightly exceeds the full cache (43.01 vs.\ 42.98), suggesting that selective merging can act as implicit attention filtering under moderate compression.

\begin{table}[H]
\centering
\begin{minipage}[t]{0.48\linewidth}
\centering
\caption{LongBench average (LLaMA-3.1-8B, 3.5k tokens) at different KV retention ratios.}
\label{tab:ratio_llama}
\scriptsize
\resizebox{\linewidth}{!}{%
\begin{tabular}{l|cccccc}
\toprule
\textbf{Method} & \textbf{10\%} & \textbf{25\%} & \textbf{50\%} & \textbf{70\%} & \textbf{80\%} & \textbf{90\%} \\
\midrule
Full Cache & \multicolumn{6}{c}{40.54} \\
\midrule
SnapKV       & 38.06 & 39.79 & 40.29 & 40.41 & 40.49 & 40.46 \\
LOOK-M       & 36.65 & 38.78 & 40.27 & 40.40 & 40.49 & 40.51 \\
SelKV(ours)  & 37.56 & 39.81 & 40.34 & 40.38 & 40.42 & 40.46 \\
\bottomrule
\end{tabular}}
\end{minipage}\hfill
\begin{minipage}[t]{0.48\linewidth}
\centering
\caption{LongBench average (Gemma-2-9B-IT, 3.5k tokens) at different KV retention ratios. SelKV leads from 25\% onward and exceeds Full Cache at 80\%.}
\label{tab:ratio_gemma}
\scriptsize
\resizebox{\linewidth}{!}{%
\begin{tabular}{l|cccccc}
\toprule
\textbf{Method} & \textbf{10\%} & \textbf{25\%} & \textbf{50\%} & \textbf{70\%} & \textbf{80\%} & \textbf{90\%} \\
\midrule
Full Cache & \multicolumn{6}{c}{42.98} \\
\midrule
SnapKV       & \textbf{41.28} & 42.16 & 42.69 & 42.87 & 42.92 & \textbf{42.96} \\
LOOK-M       & 39.11 & 41.18 & 42.37 & 42.73 & 42.84 & 42.95 \\
SelKV(ours)  & 41.05 & \textbf{42.31} & \textbf{42.80} & \textbf{42.93} & \textbf{43.01} & \textbf{42.96} \\
\bottomrule
\end{tabular}}
\end{minipage}
\end{table}

\begin{figure}[H]
\centering
\includegraphics[width=\columnwidth]{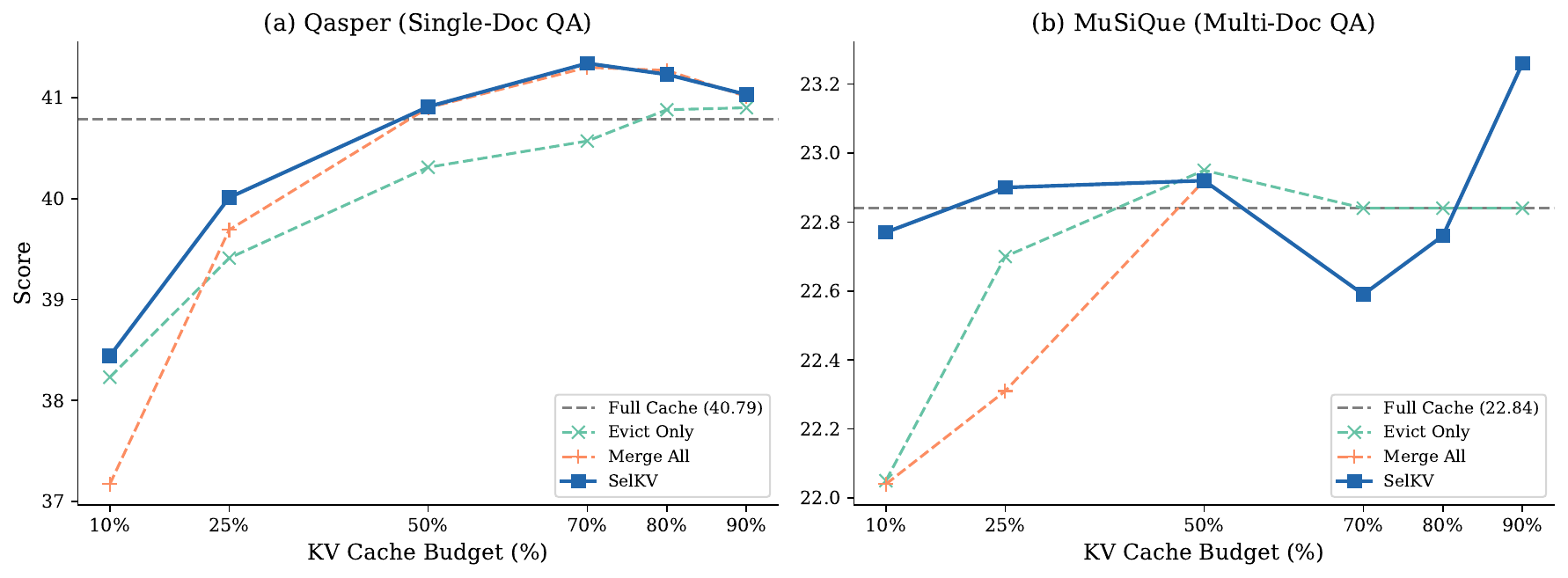}
\caption{Per-dataset ratio sweep on Gemma-2-9B-IT. Left: Qasper, where eviction outperforms blind merging but the gate protects quality by dropping dissimilar tokens. Right: MuSiQue, where merging helps and SelKV exceeds the full cache at 25\%.}
\label{fig:ratio_per_dataset}
\end{figure}

Figure~\ref{fig:ratio_per_dataset} illustrates why the adaptive gate matters by contrasting two datasets on Gemma-2. On Qasper (single-document QA), eviction outperforms blind merging at low budgets. SelKV tracks close to the better baseline at each budget and exceeds the full cache from 50\% onward, the gate adapts by dropping dissimilar tokens where merging would hurt. On MuSiQue (multi-document QA), merging helps, and SelKV at 25\% exceeds the full cache (22.90 vs.\ 22.84), demonstrating that selective merging can act as implicit attention filtering.

\section{Long-Context Evaluation (31{,}500 Tokens)}
\label{app:longcontext}

Table~\ref{tab:31500} evaluates the same four compressed methods as Table~\ref{tab:full_results} on LLaMA-3.1-8B at 31{,}500-token inputs. At this length, SelKV and SnapKV are tied and both outperform PyramidKV and LOOK-M.

\begin{table}[H]
\centering
\caption{LongBench average (LLaMA-3.1-8B, 31{,}500 tokens, 25\% KV retention).}
\label{tab:31500}
\small
\begin{tabular}{lcc}
\toprule
\textbf{Method} & \textbf{Avg (16 ds)} & $\Delta$ \\
\midrule
Full Cache    & 49.36 & -- \\
\midrule
SnapKV        & 48.46 & $-$0.90 \\
PyramidKV     & 48.26 & $-$1.10 \\
LOOK-M        & 47.16 & $-$2.20 \\
\textbf{SelKV (ours)}  & \textbf{48.46} & $-$0.90 \\
\bottomrule
\end{tabular}
\end{table}

\section{Implementation Details}
\label{app:implementation}

\paragraph{Hardware.} All experiments were run on NVIDIA H100 80GB GPUs with CUDA 12.1 and PyTorch 2.4.

\paragraph{Hyperparameters.} Unless otherwise noted, we use observation window $W = 32$, smoothing kernel $K = 5$, bucket size $B = 32$, recent window $r = 16$ tokens, compensation factor $\alpha = 0.5$, and KV retention ratio $\rho = 0.25$. For GQA models, each KV head independently selects its top-$m_s$ tokens; the final kept set is the union across heads, with any excess tokens trimmed by mean aggregate score. For MHA models (LongChat), we use per-head selection with the same observation-window procedure.

\paragraph{Evaluation.} We follow the LongBench evaluation pipeline and use greedy decoding (temperature 0, no sampling). LongChat is evaluated with 31{,}500-token inputs, whereas all other models use 3{,}500-token inputs. For code-completion tasks (LCC and RepoBench-P), we evaluate 500 samples; for all other tasks, we evaluate 200 samples.

\paragraph{Runtime.} A full 16-dataset evaluation for one model takes approximately 1-2 hours on a single H100 GPU, depending on context length. The ratio-sweep ablation (6 ratios $\times$ 16 datasets $\times$ 3 methods per model) requires approximately 8 GPU-hours per model.

\section{Generated Text Examples}
\label{app:visualization}

We show qualitative examples of generated text from LLaMA-3.1-8B-Instruct under Full Cache and SelKV (25\% KV budget). All outputs use greedy decoding (temperature~0).

\paragraph{Example 1: Qasper (Single-Document QA).}
A scientific article about characterizing political fake news on Twitter (3{,}500 tokens) is provided as context. The question asks: \emph{``How is the ground truth for fake news established?''}

\smallskip
\noindent\textbf{Ground truth:} Ground truth is not established in the paper.

\smallskip
\noindent\fbox{\parbox{0.97\columnwidth}{
\textbf{Full Cache:} \small [...] who found that fake news items are more likely to be shared than legitimate articles coming from trusted sources, and that they tend to exhibit a larger level of polarization. \textit{Conclusion:} In this paper, we have presented a preliminary characterization of fake news in Twitter by looking into meta-data embedded in tweets. Our results show that there are specific pieces of meta-data about tweets that may allow the identification of fake news. In particular, we found that tweets containing fake news are shorter-lived, have a larger proportion of friends/followers, and include more URLs [...]
}}

\smallskip
\noindent\fbox{\parbox{0.97\columnwidth}{
\textbf{SelKV (25\%):} \small [...] who found that fake news is more likely to be polarized than legit news. \textit{Conclusion:} In this paper, we have characterized political fake news on Twitter by its meta-data. We found that tweets containing fake news are more likely to be created by unverified accounts, are shorter-lived, and are more likely to be polarized. We also found that they are more likely to be spread by accounts with a larger proportion of friends/followers, and that they contain more URLs [...]
}}

\smallskip
Our method produces coherent, topically consistent summaries of the article's findings. It preserve the key conclusions (fake news is shorter-lived, more polarized, spread by accounts with more friends/followers) while differing only in phrasing.

\paragraph{Example 2: SAMSum (Dialogue Summarization).}
A dialogue dataset sample (3{,}500 tokens of few-shot examples) is provided. The target dialogue is between William and Emilia.

\smallskip
\noindent\textbf{Ground truth:} Emilia is still angry.

\smallskip
\noindent\fbox{\parbox{0.97\columnwidth}{
\textbf{Full Cache:} \small Vicki will come to the place at 21:30 tomorrow night. They will be open until 22:00. [continues generating further dialogue-summary pairs]
}}

\smallskip
\noindent\fbox{\parbox{0.97\columnwidth}{
\textbf{SelKV (25\%):} \small Vicki is asking about the store hours for tomorrow night. [continues generating further dialogue-summary pairs]
}}

\smallskip
Our method (SelKV) correctly follows the few-shot pattern (generating summary for subsequent dialogues). It produced slightly more concise summary while preserving factual accuracy.

\end{document}